\documentclass[twoside,twocolumn,10pt]{article}
\usepackage{wscg}           % includes a number of other packages (e.g., myalgorithm)
\RequirePackage{ifpdf}
\ifpdf
 \RequirePackage[pdftex]{graphicx}
 \RequirePackage[pdftex]{color}
\else
 \RequirePackage[dvips,draft]{graphicx}
 \RequirePackage[dvips]{color}
\fi

\usepackage{multirow}

\title{CNN-based Game State Detection for a Foosball Table}

\author{
\parbox{0.2\textwidth}{\centering
David Hagens\\[1mm]
ORDIX AG\\
Karl-Schurz-Stra{\ss}e 19a\\
33100, Paderborn, Germany\\[1mm]
ddh@ordix.de
}
\hspace{0.05\textwidth}
\parbox{0.2\textwidth}{\centering
Jan M. Knaup\\[1mm]
ORDIX AG\\
Karl-Schurz-Stra{\ss}e 19a\\
33100, Paderborn, Germany\\[1mm]
jkp@ordix.de
}
\hspace{0.05\textwidth}
\parbox{0.2\textwidth}{\centering
Elke Hergenr\"other\\[1mm]
Darmstadt University of Applied Sciences\\
Sch\"offerstra{\ss}e 3\\
64295, Darmstadt, Germany\\[1mm]
elke.hergenroether@h-da.de
}
\hspace{0.03\textwidth}
\parbox{0.23\textwidth}{\centering
Andreas Weinmann\\[1mm]
Algorithms for Computer Vision, Imaging and Data Analysis Group,\\
Hochschule Darmstadt, 
Germany\\[1mm]
andreas.weinmann@h-da.de
}
}

\usepackage{url}
\makeatletter
\def\Uslash{\mathbin{\mathchar`\/}\@ifnextchar{/}{\kern-.15em}{}}
\g@addto@macro\UrlSpecials{\do \/ {\Uslash}}
\def\Ucolon{\mathbin{\mathchar`:}\@ifnextchar{/}{\kern-.1em}{}}
\g@addto@macro\UrlSpecials{\do : {\Ucolon}}
\makeatother

\usepackage{color}

\begin{document}

\twocolumn[{\csname @twocolumnfalse\endcsname

\maketitle  % full width title

\begin{abstract}
\noindent
The automation of games using Deep Reinforcement Learning Strategies (DRL) is a well-known challenge in AI research. 
While for feature extraction in a video game typically the whole image is used, this is hardly practical for many real world games. 
Instead, using a smaller game state reducing the dimension of the parameter space to include essential parameters only seems to be a promising approach. In the game of Foosball, a compact and comprehensive game state description consists of the positional shifts and rotations of the figures and the position of the ball over time. In particular, velocities and accelerations can be derived from consecutive time samples of the game state.
In this paper, a figure detection system to determine the game state in Foosball is presented. 
We capture a dataset containing the rotations of the rods which were measured using accelerometers and the positional shifts were derived using traditional Computer Vision techniques (in a laboratory setting). This dataset is utilized to train  
Convolutional Neural Network (CNN) based end-to-end regression models
to predict the rotations and shifts of each rod. We present an evaluation of our system using different state-of-the-art CNNs as base architectures for the regression model. We show that our system is able to predict the game state with high accuracy. 
By providing data for both black and white teams, the presented system is intended to provide the required data for future developments of Imitation Learning techniques w.r.t. to observing human players.
\end{abstract}
\subsection*{Keywords}
Game State Detection, Computer Vision, Deep Learning, Foosball, Deep Reinforcement Learning, Imitation Learning
\vspace*{1.0\baselineskip}
}]

%%%%%%%%%%%%%%%%%%%%%%%%%%%%%%%%%%%%%%%%%%%%%%%%%%%%%%%%%%%%%%%%%%%%%%%%%%%%%

\section{Introduction}

\copyrightspace
State detection based on Computer Vision techniques has been used in the automation of games using Reinforcement Learning, cf. \cite{openai2019dota,Silver2016}. The common way is to take an input image stream, e.g. the whole screen in a video game cf. \cite{openai2019dota}, and predict a next action based on a DRL system. In contrast to video games, the usage of whole images is often not practical when automating real-world games. In this case, using a lower dimension abstraction of the game, for which we employ the term game state,
can be advantageous for the DRL training and prediction. The game of Football is a good example for the automation of complex real-world games \cite{Kitano1997}. 
A better accessible scenario is found in the game of Foosball, which provides a stable, more controllable environment, cf. \cite{weigel2005kiro,weigel2003kiro}. More concrete, the game state can be defined as the positional shift and the rotations of the figure rods plus the position of the ball as a function of time. In particular, the velocities of the figure rods and the velocity of the ball can be approximately calculated using multiple consecutive time samples. We note that, specifically in Foosball, using the described game state instead of an overall image reduces the parameter space for a DRL agent significantly. 
In order to provide the required state space data, a game state detection system is needed to extract the game state and provide the data to the DRL agent.

\begin{figure}[htb]
    \centering
    \includegraphics[width=\linewidth]{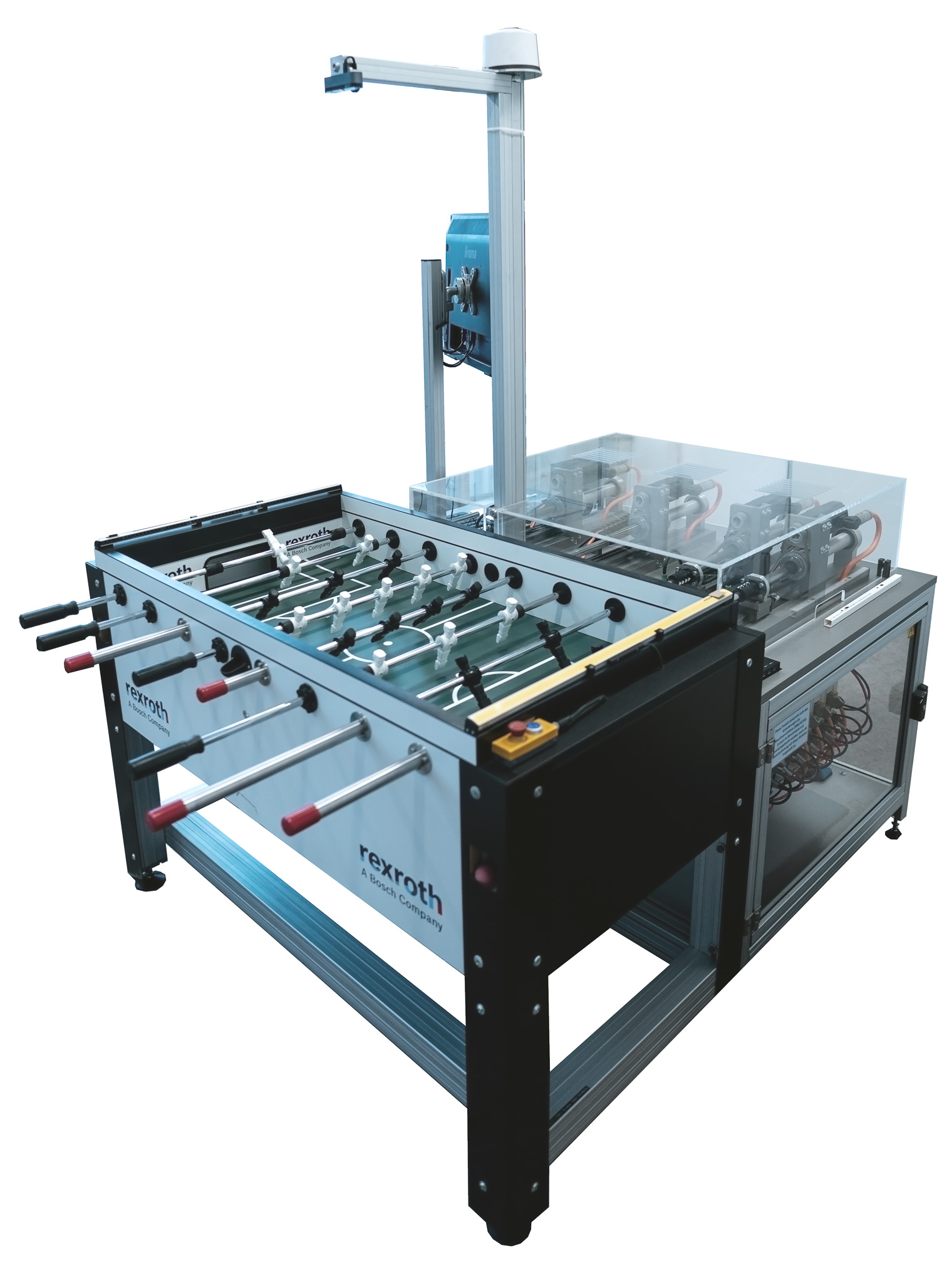}
    \caption{The physical Foosball table. The black team is controlled by industrial linear and rotary motors while the white team is controlled by humans. A Logitech BRIO webcam captures the playing field in a top down perspective.}
    \label{fig:physical_table}
\end{figure}

\begin{figure}[htb]
    \centering
    \includegraphics[width=\linewidth]{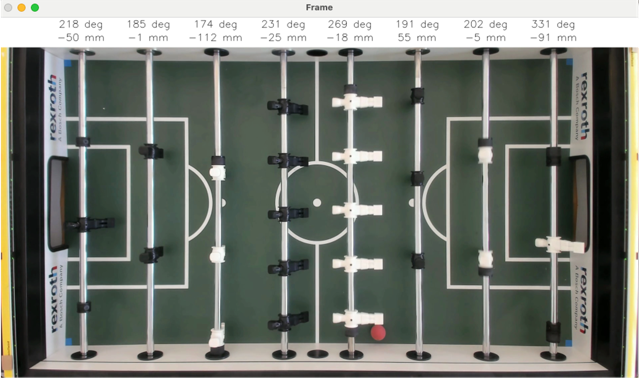}
    \caption{A working example of our figure detection system. The predicted positional shift and rotation angle of each rod is printed above the rod. The captured data can be used by a DRL agent through a ZeroMQ based data provisioning system.}
    \label{fig:working_example}
\end{figure}

In this paper, we present a game state detection system for the Foosball table shown in Fig. \ref{fig:physical_table}. The Foosball table is automated using industrial motors on the black team while the other, white team is human-controllable. At the top of the table, a Logitech BRIO webcam is mounted which captures the playing field in a top down perspective. Currently, the motors report their shifts and rotations, so the game state of the black figures is available as a potential ground truth. Additionally, 
there exists work on the detection of the position and velocity 
of the ball \cite{DeBlasi2021}. 
The game state of the white figures is unknown.  
Our system developed in this work is able to detect the game state of all figures (black and white) of the Foosball table using Deep CNN and Computer Vision. We utilize the available game state data for the black figures as a ground truth to validate our system. A working example of our system is shown in Fig. \ref{fig:working_example}. As all figures are included in the game state, our system can also be used to capture the state of non-automated Foosball tables. In this respect, we are preparing and contributing to future Imitation Learning experiments (based on matches of only human players) where the whole state space is planned to be captured by Computer Vision techniques.

The paper is structured as follows: Section 1 introduces this paper and provides a brief overview of related work. In section 2 we describe our approach to game state detection. We evaluate our system and present results in section 3. In section 4 we conclude the paper and present aspects of future work.

\subsection{Related Work}

Several studies conducted research in automating a Foosball table. The sub-process of game state detection is a key part of the automation process \cite{gutierrez2013,hernandez2019design,weigel2003kiro}. While the detection of the ball is a necessity for a rudimentary automation of a Foosball table, Enos et al. \cite{enos2012football} and Gashi et al. \cite{Gashi2023} note the importance of the detection of the figures to enable a dynamic game play. Due to the focus on the basic automation, most studies do not attempt the figure detection in their work. Bambach et al. \cite{bambach2012real} and Horst et al. \cite{Horst2024} only examined the game state detection without addressing the actual automation process. In contrast, Gashi et al. \cite{Gashi2023}, Rohrer et al. \cite{Rohrer2021} and Zhang et al. \cite{zhang2007learning} only addressed the automation by extending already existing automated Foosball tables and using the established state detection methods.

The automation Hardware is usually built around linear and rotary motors to control the rods with additional sensors to measure the shift and rotation, cf. \cite{DeBlasi2021,janssen2010real,mohebi2022study,weigel2003kiro}. Over time, the research shifted from rule based algorithms, cf. \cite{weigel2003kiro,weigel2004adaptive}, towards DRL based machine learning models, cf. \cite{DeBlasi2021,Gashi2023,Rohrer2021}. Imitation Learning methods were also used to improve the early rule based approaches \cite{zhang2007learning}. 

All above studies use a camera and computer vision techniques to detect the game state. Mohebi \cite{mohebi2022study} describes additional methods for the ball state detection, including the usage of Bluetooth, a touch sensible screen as the playing field and a grid of light emitters and detectors. Most studies, however, rely on color segmentation based approaches to detect a distinct colored ball, cf. \cite{bambach2012real,enos2012football,gutierrez2013,janssen2012ball,stefani2014automated}. The state of the figures is often not considered. Weigel et al. \cite{weigel2003kiro} implemented a rudimentary figure detection system which they discarded in their next iteration \cite{weigel2005kiro}. Bo\v{s}nak et al. \cite{bosnak2020} proposed the usage of visual marker patterns to detect the rotation of the rods. Janssen et al. \cite{janssen2010real} used a MRI scanner to detect figures in a 3D scan but only used this information to enhance the ball detection system. Mohebi \cite{mohebi2022study} proposed the usage of additional sensors for the figure detection. Other studies addressed the figure detection by using a similar color segmentation approach as already used for the ball detection, cf. \cite{aeberhard2007single,mohebi2022study}. In contrast, Horst et al. \cite{Horst2024} concentrated their work on detecting the game state of an automated Foosball table by implementing a proof-of-concept for the midfield, human-played figures. Their setup includes the utilization of the YOLOX object detector \cite{yolox} to find the individual figures and a custom regressor network to predict the rotation angle. 

\subsection{Contributions}

In this work, we present a figure detection system for the physical automated Foosball table used by Horst et al. \cite{Horst2024}, De Blasi et al. \cite{DeBlasi2021} as well as \cite{Gashi2023,Rohrer2021}, the later two references focusing on the DRL part.
Our work extends previous work of Horst et al. \cite{Horst2024} who derived a first proof-of-concept using CNN-based Computer Vision methods.

Our contributions can be summarized as follows.
\textit{(a)} We created and verified a ground truth dataset for the training of a CNN-based detection model;
\textit{(b)} We derive an improved figure detection system by basing on end-to-end learning; 
further, the new system is able to detect all white figures and additionally the black, computer controlled figures. 
\textit{(c)} In the developed end-to-end setup, we trained and evaluated multiple feature extractor backbones, including ResNet \cite{resnet}, MobileNet \cite{mobilenet} and EfficientNet \cite{efficientnet};
\textit{(d)} We propose a data provisioning system based on ZeroMQ \cite{hintjens2013zeromq}.
Concerning \textit{(a)} we use accelerometers and the motors of the Foosball table to measure the rotations of the rods. Additionally, we derive the shift by utilizing traditional CV techniques. While \cite{Horst2024} captured all data in one video using OLED screens to display the rotation of the rods, we discard the screens and OCR step by directly reading from the micro-controller, therefore reducing complexity and fixing noted issues.
Concerning \textit{(b)} we discard the YOLOX object detector \cite{yolox} which was used in \cite{Horst2024} to detect each individual figure. Instead, we aim for an end-to-end regressor network for shift and rotation detection on a per-rod basis. Additionally, we extend and refine the ideas of \cite{Horst2024} to include all figures of the Foosball table and to fix issues noted there.
Furthermore, we contribute to the DRL research started by De Blasi et al. \cite{DeBlasi2021} and extended by Rohrer et al. \cite{Rohrer2021} and Gashi et al. \cite{Gashi2023}. 
In particular, by deriving the whole state space (including the black figures), we are preparing for Imitation Learning approaches which require a system to extract the game state of pre-recorded human matches (including both black and white figures).

In contrast to other previous research, color based approaches, cf. \cite{aeberhard2007single,bambach2012real,stefani2014automated,weigel2003kiro}, are not applicable since the figures are not colored with distinguishable colors. While we also use additional sensors for creating a ground truth, a similar approach to Mohebi \cite{mohebi2022study} is not applicable due to a non-permanent hardware modification constraint. We use state of the art deep CNN for the game state detection which removes the need for a calibration step ahead of time and creates a more robust and stable system in contrast to traditional CV techniques. 

\section{Game State Detection}

In the following, we present our approach to detecting the game state of a Foosball game. 
Relying on previous work on detecting the ball \cite{DeBlasi2021}, 
we focus on detecting the figures.
More precisely, we detect the positional shifts and rotations of the figures.
This amounts to detecting the shift and rotations of the corresponding rods 
since the relative position and the angles do not change within the rod.
The velocities of the figures can be calculated through consecutive shifts and rotations. First, we describe our approach to creating a ground truth dataset using accelerometers and the motors of the Foosball table. 
Afterwards, we develop end-to-end regressor models basing on widely used CNN backbones as feature extractor networks.

\subsection{Dataset Creation}
\label{sec:dataset_creation}

The training of our end-to-end regressor network requires the presence of a dataset consisting of the shifts and rotations of all rods of our Foosball table. 
For the white, human controlled rods, the shift and rotation values are not available. In contrast, the motors which control the black rods report their state. We use the reported state as a ground truth to verify our dataset creation system.

\begin{figure}[t]%[htb]
    \centering
    \includegraphics[width=0.7 \linewidth]{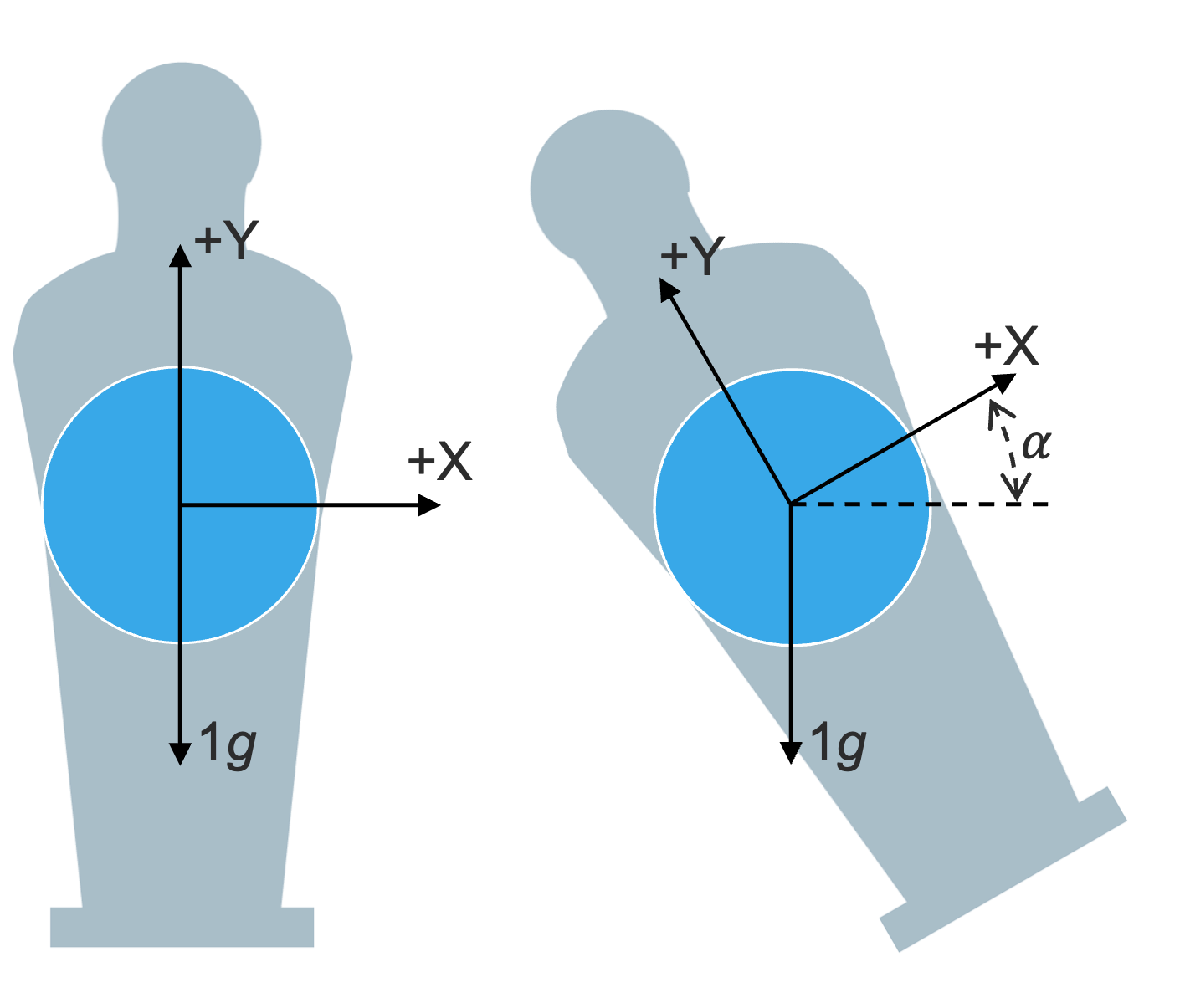}
    \caption{Measuring the rotational shift relative to the ground ($\alpha$) using a two-axis accelerometer. Since the gravitational force of $1g$ is fixed and due to the orthogonal alignment of the $X$ and $Y$ axis, the measured accelerations on those axis are proportional to the sine and cosine of $\alpha$.}
    \label{fig:two_axis_tilt_sensing}
\end{figure}

The shift of the white rods can be calculated using traditional Computer Vision techniques. In contrast, the rotation cannot be validated with traditional CV methods due to different perspectives and no ground truth being available. Therefore, our approach includes the usage of accelerometers to measure the rotation in the real world. An accelerometer measures the linear acceleration on a defined axis. When fixed to a specific point, the gravitational force is a linear acceleration which can be measured by an accelerometer. Using two orthogonal axis, the relative tilt to the ground can be calculated, cf. \cite{fisher2010using,Pham2022}. As shown in Fig. \ref{fig:two_axis_tilt_sensing}, the measured acceleration $A$ 
is related to the angle $\alpha$: 
$\frac{A_{X,OUT}}{A_{Y,OUT}}=\frac{1g\times\sin(\alpha)}{1g\times\cos(\alpha)}=\tan(\alpha)$ which results in $\alpha=\tan^{-1}(\frac{A_{X,OUT}}{A_{Y,OUT}})$ with $A_{X,OUT}$ as the measured acceleration on the X axis and $A_{Y,OUT}$ as the measured acceleration of the Y axis. The default rotation is defined as a figure standing vertically with the head up. The rotation is measured as 0 degrees in the default rotation while the motors of the black figures report 180 degrees as default. Therefore, the measured rotations are shifted to also use a default rotation at 180 degrees.

\begin{figure}[htb]
    \centering
    \includegraphics[width=\linewidth]{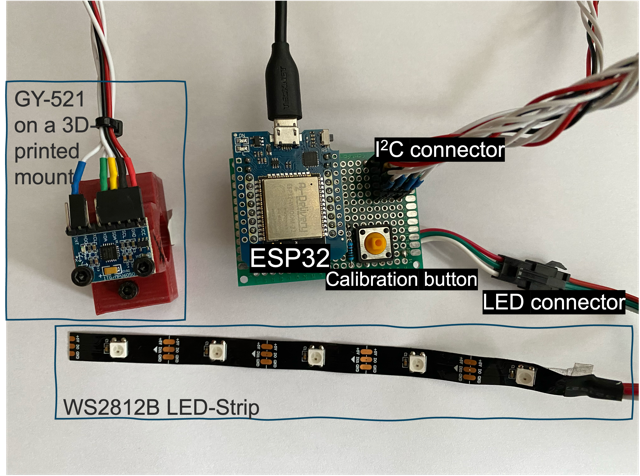}
    \caption{The hardware used to measure the rotation of the white figures. Four GY-521 modules with MNPU6050 accelerometers on 3D-printed mounts were screwed on top of the rods. The accelerometers are connected to an ESP32 based micro-controller via an $I^2C$ connector. Additionally, we included a WS2812B-based LED strip to indicate the internal timing and a button to calibrate the zero-point of the accelerometers.}
    \label{fig:hardware}
\end{figure}

We validated two different types of accelerometers, namely the MPU6050 on GY-521 breakout boards and the ADXL345. By measuring the rotation of the black rods with those sensors we could verify the measurements and estimate errors between the actual rotation and the measurements. We found that the MPU6050 sensor measured the rotation with a higher accuracy but still had a mean absolute error of around 5 degrees with outliers between 20 and 25 degrees. We also observed that some of the sensors tested had calibration errors which where visible in a systematic deviation between the real and measured rotation angle. 
To minimize the deviation between the measurements and the real rotation in our ground truth, we rejected the corresponding sensors.
Fig. \ref{fig:hardware} shows an image of our measurement hardware.

\begin{figure}[htb]
    \centering
    \includegraphics[width=\linewidth]{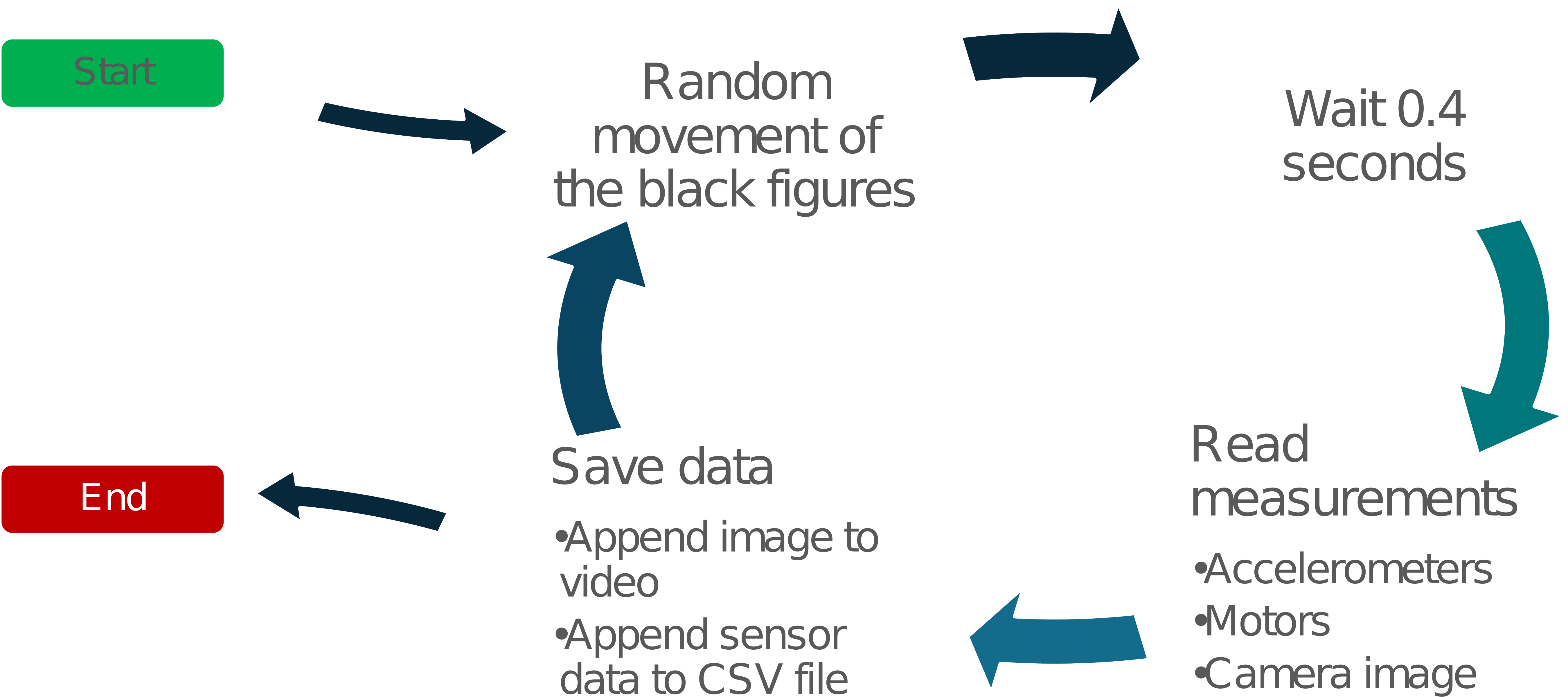}
    \caption{Our proposed dataset capturing process. One iteration of the process corresponds to one frame taken by the camera. While the black figures are moved automatically, the white figures need to be moved manually during the dataset capturing to get a versatile set of different shifts and rotations.}
    \label{fig:capturing_process}
\end{figure}

The capturing process, as illustrated in Fig. \ref{fig:capturing_process}, consists of four steps which are repeated. First, we move and rotate the black rods randomly to get diverse states. As the white rods are not connected to motors, those were moved manually by hand. Next, the system waits 0.4 seconds to let the motors finish their movements. This reduces the risk of motion blur and a faulty state reported by the motors, as they report their final position and not their current shift. Afterwards, the measurements of the motors and the accelerometers are retrieved and a snapshot of the webcam is captured. In the last step, the camera image is appended to an MP4-encoded video and the measurements are saved in a CSV file. This process is repeated $N$ times to get $N$ individual game states with a corresponding camera image in the dataset. The resulting dataset contains the positions and rotations of the black rods, the rotations of the white rods and the camera image.

\begin{figure}[htb]
    \centering
    \includegraphics[width=\linewidth]{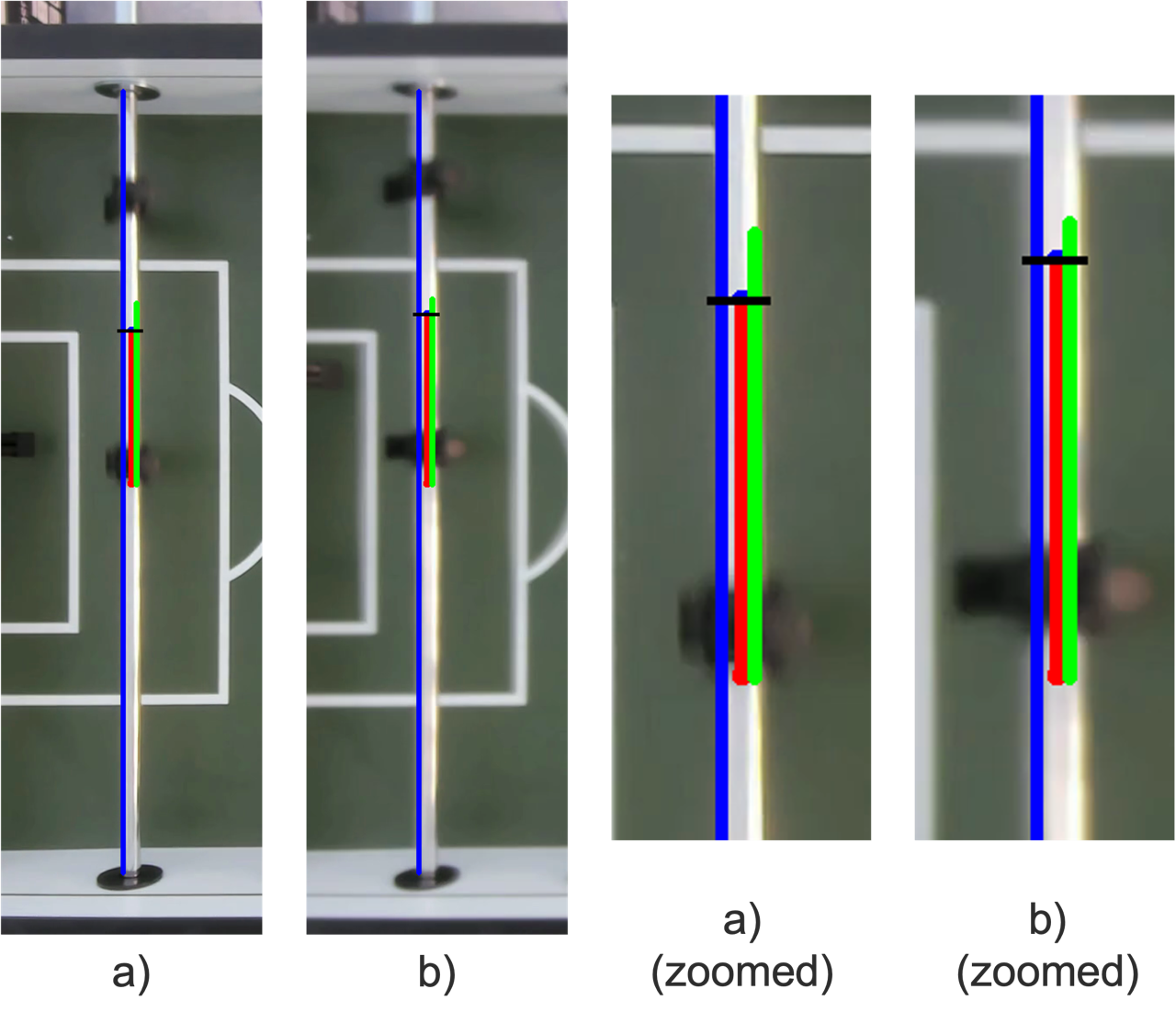}
    \caption{Outliers with a high deviation between the calculated shift (red line) and the reported shift (green line). While the motors report the final shift of the rod, regardless if the rod is still moving, the calculation is based on the current image. Therefore, the calculated shifts are more accurate.
    }
    \label{fig:deviation_outlier}
\end{figure}

For deriving the positional shifts of the white rods in the data setup we use a traditional Computer Vision approach. We start by defining a one pixel wide column in the center of each rod and applying a binary thresholding. We search for connected groups of black pixels in this column to find the rubber stoppers which are mounted at each end of the rods. Using those stoppers we can find the center of each rod. We calculate the actual shift by converting the offset between the center of the table and the center of the rod from pixels to mm.
We validated this approach by calculating the shifts of the black rods. We observed, that the reported shift sometimes differs from our calculations. As illustrated in Fig. \ref{fig:deviation_outlier}, our calculations were more accurate. We conclude, that the rods were still moving when the image was taken resulting in an offset between the reported and the actual shift. The observed motion blur pattern, as shown in Fig. \ref{fig:deviation_outlier}, confirms this assumption. While the images are not blurred, the rods contain a vertical motion blur but no horizontal bluring. Since this motion blur will also be present when detecting the game state of a real Foosball game, we keep the faulty images in the dataset but use our calculated shift as our ground truth.

\begin{figure}[htb]
    \centering
    \includegraphics[width=\linewidth]{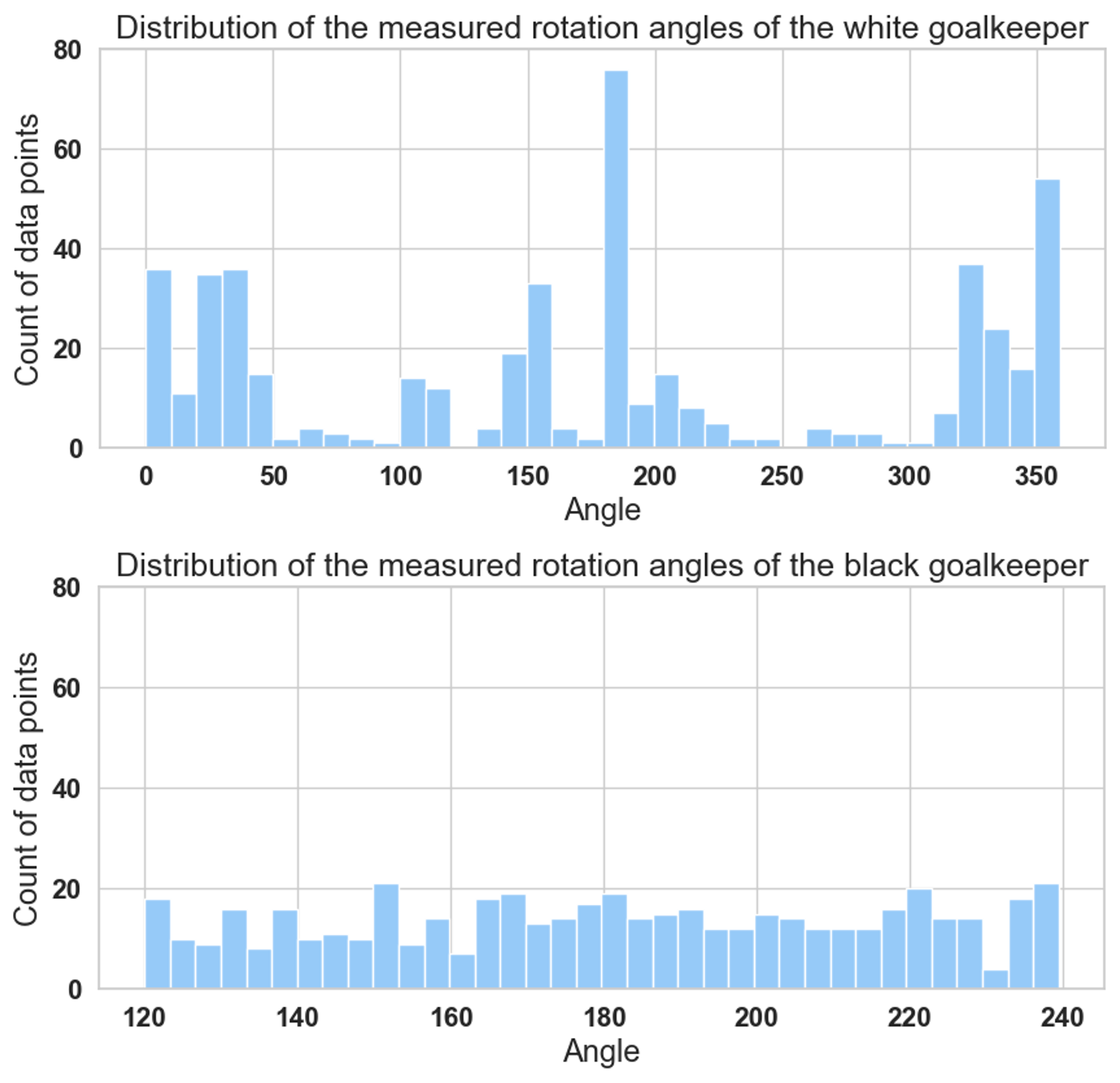}
    \caption{Distribution of the rotation of the black and white goalkeepers. The black goalkeeper was moved randomly by the computer which results in a uniform distribution of measured rotation angles. In contrast, the white goalkeeper was moved manually by hand resulting in a non-uniform distribution with a bias to specific angles. This effect is also present on the other white rods.
    }
    \label{fig:goal_rotation_dist}
\end{figure}

Summing up, the derived data set consists of 500 images containing the corresponding shifts and rotations of the black and white figures. Due to our automated data capturing process using the accelerometers for the measurement of the rotations and an automated CV process for the shift calculation, we were able to avoid a manual labeling process. We note however, that in some occasions, minor manual adaptions were necessary to accommodate small deviations in the camera angle. The positions and rotations of the black figures are uniformly distributed. Due to the manual movement of the white rods, the positions and rotations could not reach a uniform distribution and include some bias towards specific values, c.f. Fig. \ref{fig:goal_rotation_dist}. Albeit not attempting to detect the ball, we included it in the dataset to get more realistic images. 

\subsection{End-to-End Regressor Networks} \label{sec:subsEnd2End}

For the game state detection on our Foosball table, we propose the utilization of common state-of-the-art Image Classification models like ResNet \cite{resnet}, EfficientNet \cite{efficientnet} and MobileNet \cite{mobilenet}
as a backbone for feature extraction. 
As the shifts and rotations of the rods are continuous numerical variables, 
the problem at hand constitutes a regression problem. 
Thus, classification models/networks which typically consist of a feature extractor backbone and a classification head cannot be used directly.
However, we may reuse the feature extractor (cf. \cite{Aloysius2017}; which mostly consists of convolutional layers) and replace the classification head with a customized regression head.
This approach fits within the idea of fine-tuning on a downstream task using parts of a neural network trained for another task on a large set of training data; in particular, we here use already proven feature extractor architectures trained for classification on ImageNet \cite{imagenet}.

To account for the periodicity of angles, we decided to use 
$\mathrm e^{i \varphi} = (\cos \varphi, \sin \varphi)$ as variable for regression, i.e.,
we predict
the sine and cosine of the angles. Additionally, the shift is scaled to a range of -1 to 1 
which balances the influence of the shift, the sine and the cosine.
Ultimately, as a regression head, we used a linear layer with an output dimension of three and no activation function to predict the values $\left(s,\cos \varphi, \sin \varphi\right)$
where the symbol $s$ denotes the shift. 

 To mitigate the problem of perspective distortion, which occurs due to the wide field-of-view of the camera, we use an individual regressor model per rod, resulting in 8 models which are trained and inferred sequentially. Each regressor uses a pre-defined cutout of the overall camera image. We custom-trained the 8 models using four different base architectures from ResNet18, ResNet50 \cite{resnet}, EfficientNetV2 \cite{efficientnet} and MobileNetV3 \cite{mobilenet} as feature extractors. The weights of the feature extractors are initialized using transfer learning with the provided ImageNet1K \cite{imagenet} weights of the base models from PyTorch. The custom regression head is initialized randomly. Each regressor is trained over 50 epochs using the MSE loss function and the Adam optimizer \cite{kingma2017adam} with a fixed learning rate of 0.001. During the training process, we used 80 \% of our dataset (400 images) to train the CNNs and 20 \% (100 images) for validation.

\subsection{Data Provisioning System}

As our work contributes to the automation of our Foosball table using DRL or Imitation Learning techniques, a data provisioning system needs to be implemented which should \textit{(a)} use standardized formats; \textit{(b)} be easily accessible with a minimal need of further Hard- or Software; \textit{(c)} not introduce high latency into the system. We propose a publish-subscribe messaging system based on ZeroMQ \cite{hintjens2013zeromq} which satisfies those requirements.

\begin{figure}[htb]
    \centering
    \includegraphics[width=\linewidth]{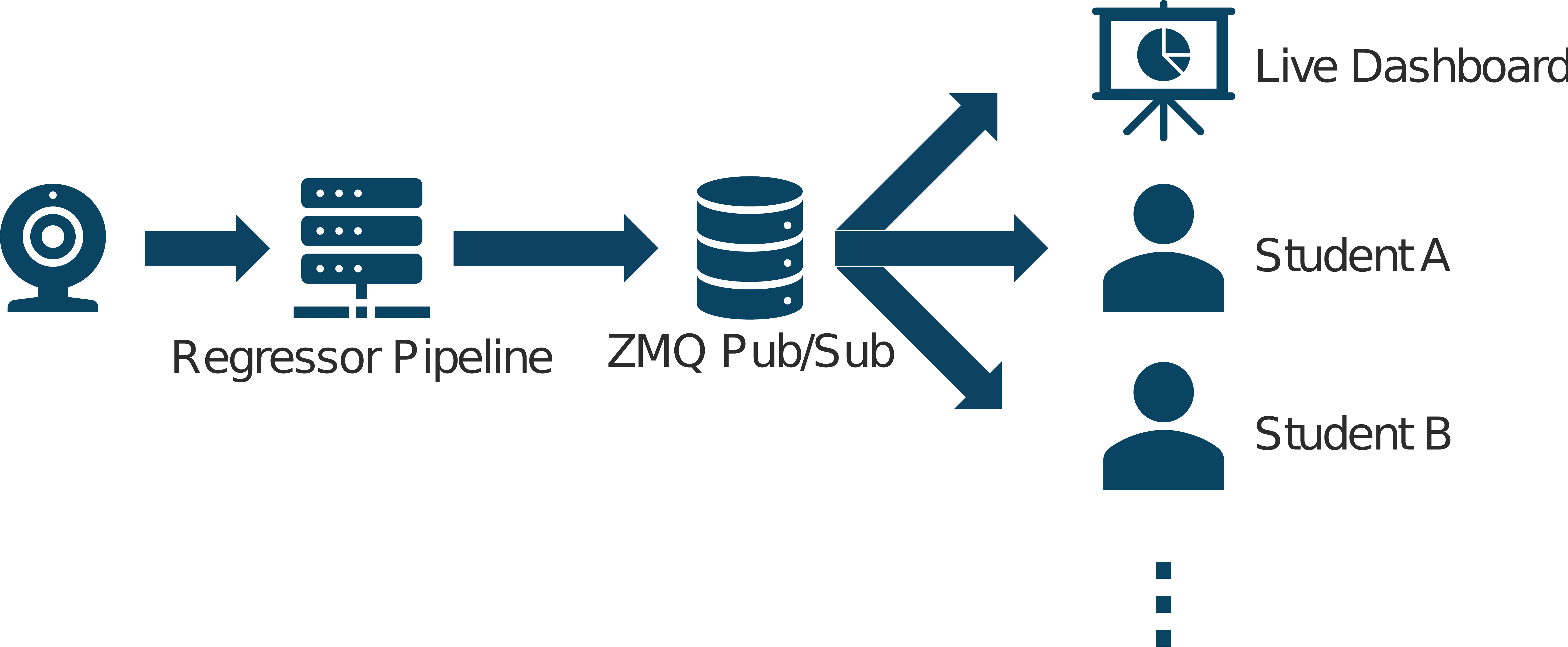}
    \caption{The proposed game state detection pipeline including the data provisioning system. Due to the utilization of ZeroMQ, multiple clients can connect and receive the game state data simultaneously.}
    \label{fig:pipeline}
\end{figure}

\begin{table*}[t]
\label{tab:position_feature_extractor}
\resizebox{\linewidth}{!}{%
\begin{tabular}{c|cccc|cccc|c}
\textbf{Feature Extractor} & \multicolumn{4}{c|}{\textbf{Black Rods}} & \multicolumn{4}{c|}{\textbf{White Rods}} & \textbf{Average} \\
                   & Goal & Defense & Midfield & Striker & Goal & Defense & Midfield & Striker &         \\ \hline
ResNet18           & 2.94 & 4.02    & 1.23     & 1.6     & 2.94 & 7.42    & 2.2      & 8.68    & 3.88    \\
ResNet50           & 2.4  & 4.02    & 1.35     & 2.77    & 4.5  & 6.49    & 1.91     & 7.83    & 3.91    \\
MobileNetV3        & 7.68 & 9.35    & 3.01     & 7.32    & 9.06 & 11.94   & 3.89     & 6.78    & 7.38    \\
EfficientNetV2     & 9.82 & 5.97    & 2.06     & 3.75    & 5.5  & 12.73   & 4.47     & 8.53    & 6.6     \\ 
\end{tabular}
}
\caption{Mean absolute error of the predicted shift in mm per rod and feature extractor.}
\end{table*}

\begin{table*}[t]
\vspace{1cm}
\resizebox{\linewidth}{!}{%
\begin{tabular}{c|cccc|cccc|c}
\label{tab:mae_feature_extractor}
\textbf{Feature Extractor}  & \multicolumn{4}{c|}{\textbf{Black Rods}}     & \multicolumn{4}{c}{\textbf{White Rods}} & \textbf{Average}       \\
                   & Goal & Defense & Midfield & Striker & Goal  & Defense & Midfield & Striker &         \\ \hline
ResNet18           & 1.23 & 1.47    & 0.88     & 1.38    & 12.64 & 13.93   & 4.96     & 10.93   & 5.93    \\
ResNet50           & 1.34 & 1.43    & 0.97     & 1.33    & 8.33  & 5.44    & 4.06     & 9.91    & 4.10    \\
MobileNetV3        & 3.46 & 2.18    & 1.72     & 4.35    & 17.86 & 22.96   & 7.69     & 21.21   & 10.18   \\
EfficientNetV2     & 6.31 & 2.14    & 2.29     & 1.69    & 14.26 & 25.32   & 13.59    & 28.68   & 11.79   \\
\end{tabular}
}
\caption{Mean absolute error of the predicted rotation angle in degrees per rod and feature extractor.}
\end{table*}

The data producer side of the pipeline, in which the shift and rotation is predicted, publishes the JSON-formatted results through a ZeroMQ TCP socket. A client can then connect to this socket, subscribe to the messages and decode the JSON data. An illustration of the data provisioning system with multiple example clients is shown in Fig. \ref{fig:pipeline}.
While this approach introduces some latency, it also has other major advantages. Firstly, multiple clients can connect to the socket simultaneously. Secondly, the TCP-based socket can also be shared through a network. This enables the possibility of the permanent installation of a dedicated game state detection server accessible to researchers. This would free resources for possible DRL experiments, as the game state detection must not be executed on the same hardware. The option of multiple connected clients allows for different simultaneous experiments as long as the actual Foosball table hardware is not required.

\section{Evaluation}

As we utilized multiple feature extractor backbones for our regression model, we evaluate the performance of the overall system per feature extractor based on the prediction accuracy measured as Mean Absolute Error (MAE) and the inference time. We define the following criteria: 

\begin{itemize}
\item[a)] The MAE for the shift should not exceed 11 mm. The feet of the figure is 22 mm wide, so an error in the positional shift of $\pm$ 11 mm would still result in a straight shot. Since every rod has different shift limits, the percentage range per rod varies between $\pm$ 8.5 \% for the defender and $\pm$ 20 \% for the midfield rod.
\item[b)] The MAE for the rotation should not exceed 42 degrees. A figure is able to block a shot if the feet of the figure are below the highest point of the ball. This is given at a maximum rotation of $\pm$ 47 degrees. Due to small rotations of the figure on impact with the ball, we deduct an offset of 5 degrees. Therefore, a figure which is predicted to stand vertically would be able to stop the ball up to an error of $\pm$ 42 degrees.
\item[c)] The inference time should be lower than 16.6 ms. Since our present camera is only able to record at 60 FPS, this is the limit which we consider to be real-time. Ultimately, the system should be as fast as possible to provide as much information to the DRL systems as possible. A shorter inference time is therefore generally preferred.
\end{itemize}

\subsection{Quantitative Evaluation}

\textbf{Evaluation of the Prediction Quality.} 
We compare the results using a ResNet18, a ResNet50 \cite{resnet}, an EfficientNetV2 \cite{efficientnet} and a MobileNetV3 \cite{mobilenet} where we use an individual regressor model per rod as detailed in Section~\ref{sec:subsEnd2End}.

Table~1
%~\ref{tab:position_feature_extractor} 
shows our evaluation results for the shift detection. 
Considering the MAE averaged w.r.t.\ all rods, all feature extractor backbones passed the defined objective -- they were below the predefined thresholds above. 
The shift detection is more accurate on rods which have a smaller movement range (e.g. the midfield rod) than on rods with a high movement range (e.g. the defense rod). 
Further, the predictions for the white rods have a higher accuracy than those for the black rods.
We believe that a possible explanation for this finding can be that,
in the training data, the shifts of the black rods follow a uniform distribution while the white rods (which were moved manually by hand) are distributed non-uniformly. 
Examining each rod individually, 
both regression networks based on the RestNet18 and on the ResNet50,
stay within the predefined error tolerance,
whereas the networks based on the MobileNetV3 and EfficientNetV2 backbones exceed the requirement on the white defender. 
Further, the ResNet18 based model performed slightly better on average than the  
ResNet50 based model. In our opinion, these observations indicates that further training data is necessary to use the full potential of deeper networks. 

In Table~2, 
%\ref{tab:mae_feature_extractor}, 
we provide the mean absolute errors of the rotation angle prediction. 
Overall, all models achieved the required maximum average error of $\pm$ 42 degrees 
having large margins to the defined threshold.
Similar to the shift prediction, the estimate for the white rods had higher MAEs compared to the black rods. In connection with this, we observe that, in addition to the non-uniform sampling, the rotation angle of the black rods is restricted to the range between 120 and 240 degrees (motor constrains) while the white rods can rotate freely. Therefore, the white rods have a higher range of possible values which can also explain the worse prediction accuracy. Furthermore, the ground truth data contains a mean error of $\pm$ 5 degrees due to the measurements with the accelerometers. As already observed in the shift detection, the ResNet based models yield lower MAE values compared to the MobileNetV3 and EfficientNetV2 based models. 

\textbf{Evaluation of the Inference Time.} 
The inference times of the models were evaluated on four different systems: 
\textit{System A:} Apple MacBook Pro 2018 with Intel Core i7 processor and AMD Radeon Pro 560X GPU; 
\textit{System B:} Apple MacBook Pro 2021 with Apple M1 Pro processor; 
\textit{System C:} PC with AMD Ryzen 9 5900X processor and NVIDIA RTX 3080 GPU; 
\textit{System D:} Cloud VM with AMD EPYC-Milan processor and NVIDIA A100 80G PCIe GPU. 
All systems use the GPUs, either through the Metal Performance Shader (MPS) backend or through NVIDIA CUDA. The models were trained using PyTorch 2.1.0 and Torchvision 0.16.0 in a Python 3.9 environment.

\begin{table*}[]
\label{tab:eval_backbones}
%\resizebox{\linewidth}{!}{%
\begin{tabular}{c|c|cc|cc|cc}
\multirow{2}{*}{\textbf{Backbone}} &
  \multirow{2}{*}{\textbf{System}} &
  \multicolumn{2}{c|}{\textbf{Inference Time (ms)}} &
  \multicolumn{2}{c|}{\textbf{Inf. per Rod (ms)}} &
  \multicolumn{2}{c}{\textbf{FPS}} \\
                                &                           & Mean    & Median  & Mean   & Median & Mean  & Median \\ \hline
\multirow{4}{*}{ResNet18}       & MacBook Pro 2018 (Sys. A) & 288.51  & 286.27  & 36.06  & 35.78  & 3.47  & 3.49   \\
                                & MacBook Pro 2021 (Sys. B) & 108.03  & 107.11  & 13.50  & 13.38  & 9.26  & 9.34   \\
                                & Gaming PC (Sys. C)        & 88.88   & 86.18   & 11.11  & 10.77  & 11.25 & 11.60  \\
                                & Cloud VM (Sys. D)         & 126.23  & 106.94  & 15.78  & 13.36  & 7.92  & 9.35   \\ \hline
\multirow{4}{*}{ResNet50}       & MacBook Pro 2018 (Sys. A) & 581.58  & 573.85  & 72.69  & 71.72  & 1.72  & 1.74   \\
                                & MacBook Pro 2021 (Sys. B) & 182.79  & 182.45  & 22.84  & 22.80  & 5.47  & 5.48   \\
                                & Gaming PC (Sys. C)        & 120.83  & 119.32  & 15.10  & 14.91  & 8.28  & 8.38   \\
                                & Cloud VM (Sys. D)         & 130.77  & 114.60  & 16.34  & 14.32  & 7.65  & 8.73   \\ \hline
\multirow{4}{*}{MobileNetV3}    & MacBook Pro 2018 (Sys. A) & 497.21  & 501.49  & 62.14  & 62.58  & 2.01  & 1.99   \\
                                & MacBook Pro 2021 (Sys. B) & 156.76  & 156.44  & 19.59  & 19.55  & 6.38  & 6.39   \\
                                & Gaming PC (Sys. C)        & 117.00  & 115.93  & 14.62  & 14.29  & 8.55  & 8.63   \\
                                & Cloud VM (Sys. D)         & 122.81  & 105.14  & 15.35  & 13.14  & 8.14  & 9.51   \\ \hline
\multirow{4}{*}{EfficientNetV2} & MacBook Pro 2018 (Sys. A) & 1538.96 & 1496.68 & 192.36 & 187.08 & 0.65  & 0.67   \\
                                & MacBook Pro 2021 (Sys. B) & 402.34  & 400.27  & 50.29  & 50.03  & 2.49  & 2.50   \\
                                & Gaming PC (Sys. C)        & 249.31  & 245.97  & 31.16  & 30.74  & 4.01  & 4.07   \\
                                & Cloud VM (Sys. D)         & 235.70  & 215.24  & 29.46  & 26.90  & 4.24  & 4.65  
\end{tabular}
%}
\caption{Mean and Median inference time for different feature extractor backbones on the different systems. All systems utilized GPU acceleration through either NVIDIA CUDA or Apple MPS. While the Cloud VM (system D) has the highest theoretical GPU performance, not all backbones could benefit from this. Instead, the lower performing RTX 3080 from system C resulted in a shorter median inference time for the small ResNet18 based model and almost equal times for the ResNet50 and MobileNetV3 based models. All systems except the cloud VM showed only small deviations between the mean and median inference time.}
\end{table*}

Table~3 %\ref{tab:eval_backbones} 
summarizes the mean and median inference times on the different systems. The inference times are measured as overall inference time and the inference time per rod. The overall time should roughly be about 8 times the inference time per rod, as all rods are inferred sequentially. Additionally, the corresponding FPS are calculated. Overall, the ResNet18 based model performed the best with the lowest inference time of 86.18 ms median on system C, thus achieving only 11.6 FPS which is significantly lower than the desired 60 FPS. The other evaluated CNN architectures could not achieve at least 10 FPS on average. The slightly better prediction quality of the ResNet50 based regressor is -in our opinion- not enough to justify the higher inference time of an additional 33.14 ms on the same hardware compared to the ResNet18 based model. If the sequential execution would be parallelized, the ResNet18 based model would probably achieve the desired 60 FPS considering the median inference time per rod of 10.77 ms on system C. The ARM-based MacBook (system B) would then also achieve the goal with 13.38 ms median inference time. 

Albeit providing a higher performing GPU, system D could not reach a better performance compared to the other systems. Our conjecture is that the system is CPU-bottle-necked due to the sequential execution of the rods. We observed an average GPU utilization of 40 \% on system C and 7 \% on system D which supports this assumption. While all systems except the cloud VM showed only small deviations of 0 to 3 \% between the median and mean inference times, system D resulted in high deviations of 9.51 \% for the EfficientNetV2 based model and 14 to 18 \% for the other models. The outliers in the inference times could indicate lower performing storage resulting in more time needed to move data between RAM and the GPU.

\subsection{Qualitative Evaluation}

In a qualitative evaluation and live tests on the Foosball table, we observed overall accurate results, e.g. illustrated in Fig. \ref{fig:working_example}. 
There were several issues, in particular concerning lightning conditions and blur which we discuss next. 

\begin{figure}[htb]
    \centering
    \includegraphics[width=\linewidth]{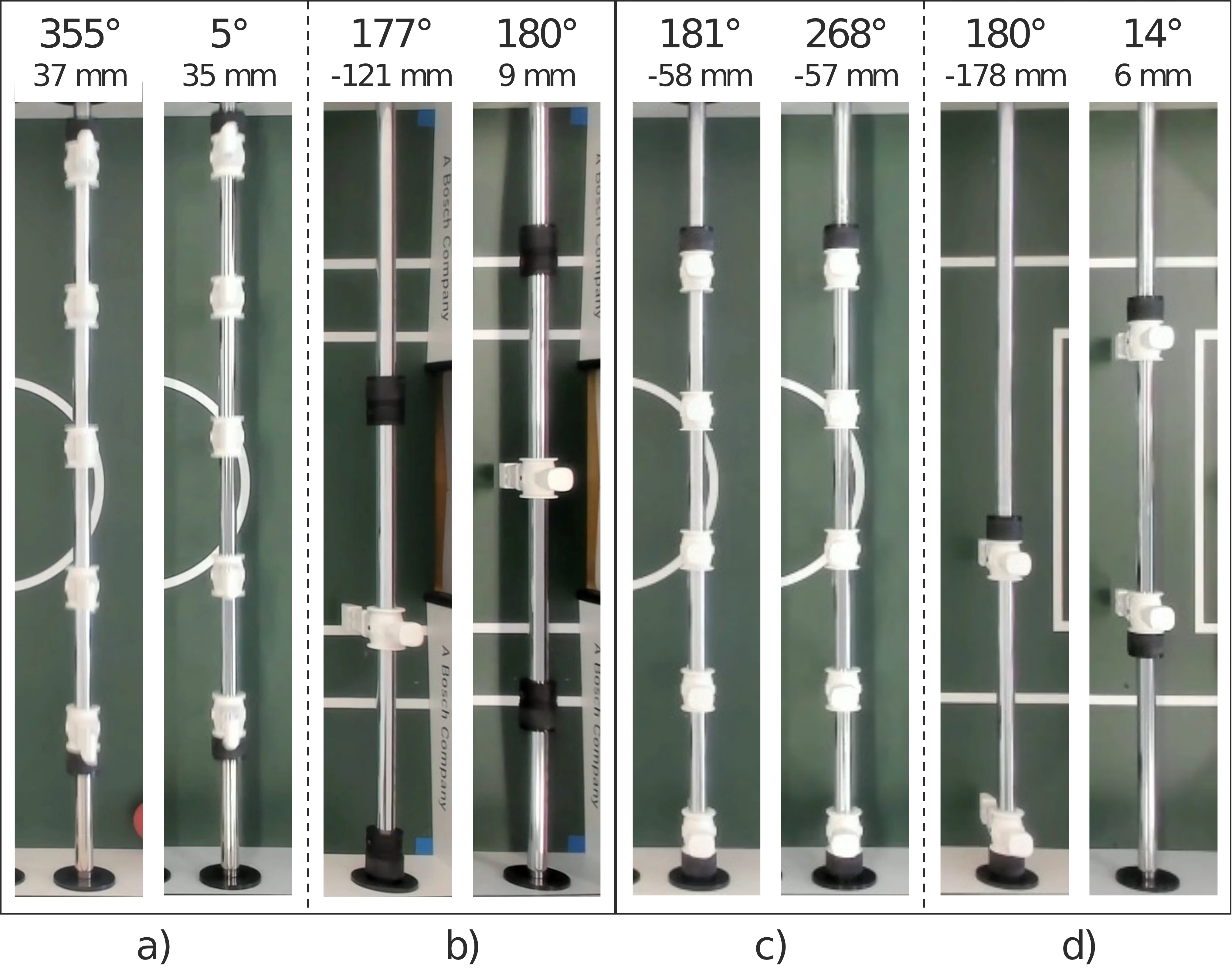}
    \caption{Examples for the influence of lighting conditions on the prediction accuracy. Each example is shown with natural light conditions (the same as in the training data) on the left and artificial light conditions on the right with a similar rotation angle. The artificial lighting results in harder shadows and a brighter background compared to the natural light. While the images in \textit{a)} and \textit{b)} predict the correct rotation in both lighting conditions, the examples in \textit{c)} and \textit{d)} show extreme examples of a deviation from the actual rotation in the artificial lighting conditions. In all cases, the position is predicted within the given boundaries.}
    \label{fig:influence_of_lighting}
\end{figure}

\textbf{Dependence of the Prediction Quality on the Lightning Conditions.} 
As illustrated in Fig. \ref{fig:influence_of_lighting}, the prediction accuracy is highly dependent on the lighting conditions which cannot be controlled. Our training data was captured with natural light in the summer while we tested the system in the winter with predominantly artificial light. As seen in Fig. \ref{fig:influence_of_lighting}, the artificial light (in the right images of each example) results in harder shadows with a brighter playing field, especially at the edges of the field. In contrast, the natural light showed only soft shadows. A more diverse dataset with different lighting conditions including natural and artificial light would improve the consistency of the system and provide a more generalized prediction model.

\begin{figure}[htb]
    \centering
    \includegraphics[width=\linewidth]{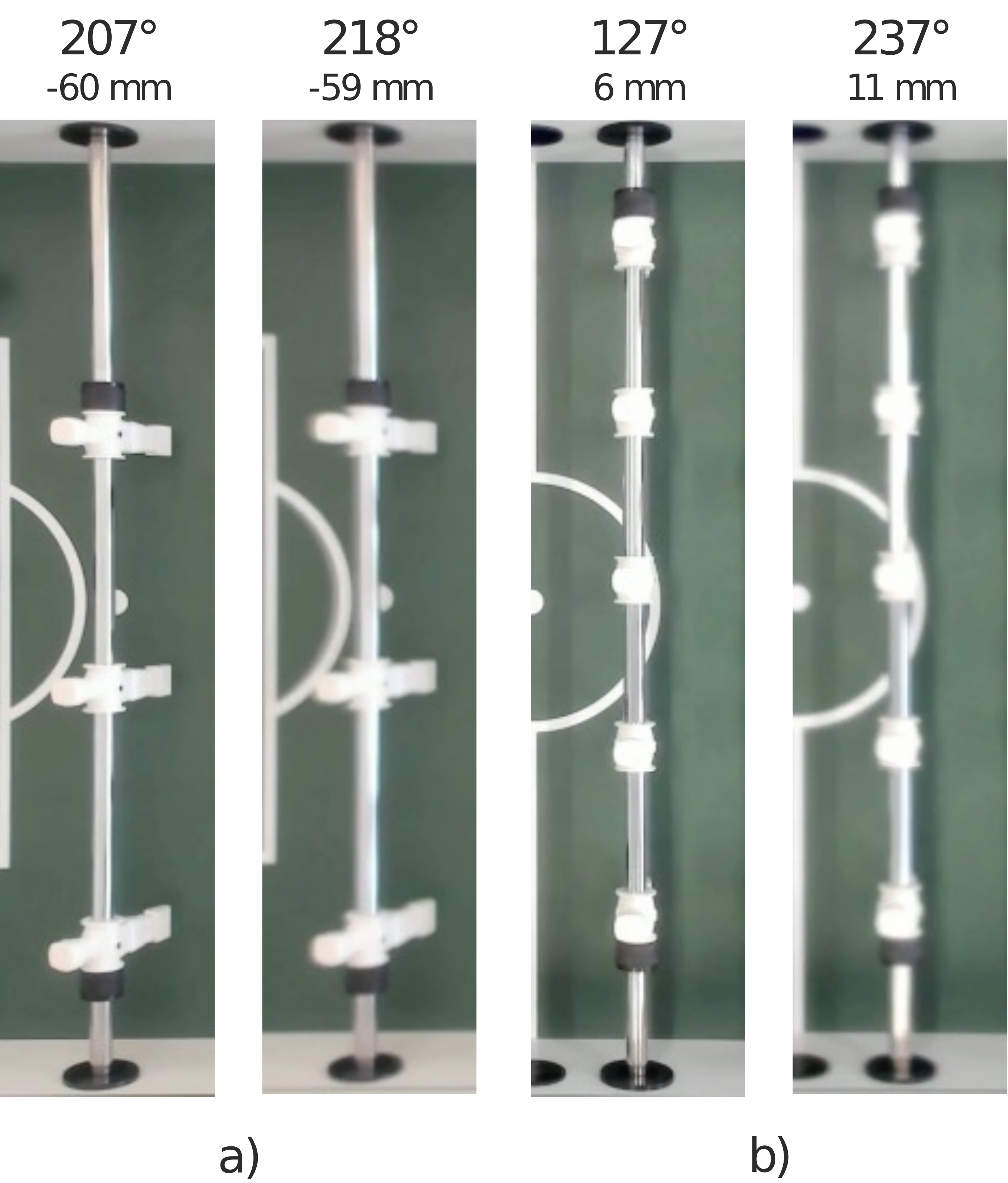}
    \caption{Two extreme examples for the influence of blurred images on the prediction quality. The left images of \textit{a)} and \textit{b)} are non blurred versions with the same positions and rotations as in the right images. The blurred images are a result of the camera performing a re-focusing of the whole image. We observed this effect to occur every 30-60 seconds.
    }
    \label{fig:blur_vs_non_blur}
\end{figure}

\textbf{Dependence of the Prediction Quality on Blur.}
We observed an influence of blur in the images on the prediction accuracy.
Since we use a standard webcam without any modifications, the exposure time and focus cannot be fixed. Commonly, a webcam uses a variable exposure time to control the brightness of the image, as the aperture is fixed. Therefore, in a darker scenario, the webcam uses a longer exposure time to get a bright image. The long exposure time can, however, result in motion blur especially on the fast moving figures of the Foosball table. Additionally, the variable focus leads to an automatic re-focusing of the camera which results in overall blurred images. In Fig. \ref{fig:blur_vs_non_blur}, the predicted rotation on the blurred, right images deviates about 11 degrees in \textit{a)} and 20 degrees in \textit{b)} compared with the same physical rotation in the non blurred counterparts in the left images.

Additionally, we observed a high influence of occluded figures and / or general unknown items in the images, e.g. by placing a hand inside the camera frame. This risk cannot be eliminated and should be addressed in the future. 

\section{Conclusion, Discussion and Future Work}

In this paper, we presented a CNN based figure detection system for a semi-automatic Foosball table: the black team was controlled by motors and the white team was controlled by human players. More precisely, we first created and verified a ground truth dataset for training CNN-based detection models. 
Then, we have derived an improved figure detection system by basing on end-to-end learning.
This contrasts the previous work \cite{Horst2024} where an intermediate object detection step utilizing a YOLO detector was used.  
Further, we included the detection of all white figures and additionally the black figures in the detection system. 
We trained and evaluated our approach using various feature extractor backbones, including ResNets, MobileNets and EfficientNets. Finally, we proposed a data provisioning system based on ZeroMQ.

We demonstrated that our system is able to detect the shifts and rotations of all figures based on a camera image in a top-down perspective. The shift and rotation detection of our system satisfy our defined requirements for all figures. Using the ResNet18-based model, we achieved mean absolute errors of 3.88 mm for the position and 5.93 degrees for the rotations of all figures. 

While the overall accuracy of the system satisfies the requirements we defined, some limitations remain. First, concerning stability, 
we observed that the lighting conditions and image blur can have significant influence on the prediction accuracy.
Second, the present system does not achieve a game state detection in real-time, i.e. at 60 FPS. 
Means to address these issues could be as follows:
Concerning speed, one possible solution is the parallelization of the regression models which are currently inferred sequentially. As shown in this paper, the inference time for one rod would achieve a real-time detection at 60 FPS. 
Concerning the stability of the system,
one potential approach is the modification of the hardware of the Foosball table. It appears to be most promising to switch to another, manually controllable camera, as the current webcam shows clear drawbacks.

A further limitation is found in the Foosball table itself. The motors controlling the black rods are subject to a rotation limitation between 120 and 240 degrees on the driver level. This safety feature cannot be overwritten without modifying the hardware and should remain to reduce the maximum velocity of the ball. A manually played Foosball table does not have this issue. The same approach as described in Sec. \ref{sec:dataset_creation} could be applied to generate 
a corresponding dataset including full 360 degree rotational movement of the black rods.

One aspect of future research is to further develop the system w.r.t. real-time capabilities and robustness such that it can be directly used in the automation process, i.e., provide the necessary information on the game state to the future RL agent controlling the non-human player in a robust way in real-time. Another line of future research is to employ (developments of) the proposed system for Imitation Learning. To this end, we plan to employ our game state detection system for capturing real Foosball games played by humans. This would improve the training of a DRL agent by reducing the need for On-Policy and / or simulation data. We note that for the creation of an Imitation Learning dataset, real-time detection is not necessary. 

% ---- Bibliography ----

\bibliographystyle{plain}
\footnotesize
\bibliography{biblio}

\begin{thebibliography}{10}

\bibitem{aeberhard2007single}
Michael Aeberhard, Shane Connelly, Evan Tarr, and Nardis Walker.
\newblock Single player foosball table with an autonomous opponent.
\newblock {\em Georgia Tech, Elect. and Comp. Engineering}, 2007.

\bibitem{Aloysius2017}
Neena Aloysius and M.~Geetha.
\newblock A review on deep convolutional neural networks.
\newblock In {\em 2017 International Conference on Communication and Signal Processing (ICCSP)}, pages 0588--0592, 2017.

\bibitem{bambach2012real}
Sven Bambach and Stefan Lee.
\newblock Real-time foosball game state tracking.
\newblock Technical report, School of Informatics and Computing, Indiana University, 2012.

\bibitem{bosnak2020}
Matev\u{z} Bo\u{s}nak and Gregor Klan\u{c}ar.
\newblock Fast and reliable alternative to encoder-based measurements of multiple 2-dof rotary-linear transformable objects using a network of image sensors with application to table football.
\newblock {\em Sensors}, 20(12), 2020.

\bibitem{DeBlasi2021}
Stefano De~Blasi, Sebastian Kl{\"o}ser, Arne M{\"u}ller, Robin Reuben, Fabian Sturm, and Timo Zerrer.
\newblock Kicker: An industrial drive and control foosball system automated with deep reinforcement learning.
\newblock {\em Journal of Intelligent {\&} Robotic Systems}, 102(1):20, 2021.

\bibitem{imagenet}
Jia Deng, Wei Dong, Richard Socher, Li-Jia Li, Kai Li, and Li~Fei-Fei.
\newblock Imagenet: A large-scale hierarchical image database.
\newblock In {\em 2009 IEEE Conference on Computer Vision and Pattern Recognition}, pages 248--255, 2009.

\bibitem{enos2012football}
Nathaniel Enos, Patrick Fenelon, Skyler Goodell, and Nicholas Phillips.
\newblock Football operator and optical soccer engine (foose).
\newblock {\em University of Central Florida, Department of Electrical and Computer Engineering}, 2012.

\bibitem{fisher2010using}
Christopher~J Fisher.
\newblock Using an accelerometer for inclination sensing.
\newblock {\em AN-1057, Application note, Analog Devices}, pages 1--8, 2010.

\bibitem{Gashi2023}
Adriatik Gashi, Elke Hergenr{\"o}ther, and Gunter Grieser.
\newblock Efficient training of foosball agents using multi-agent competition.
\newblock In Kohei Arai, editor, {\em Intelligent Computing. SAI 2023}, pages 472--492, Cham, 2023. Springer Nature Switzerland.

\bibitem{yolox}
Zheng Ge, Songtao Liu, Feng Wang, Zeming Li, and Jian Sun.
\newblock Yolox: Exceeding yolo series in 2021.
\newblock {\em arXiv}, 2021.

\bibitem{gutierrez2013}
Juan~David Gutierrez-Franco, John Inlow, Jesse Graham, and Larry Huang.
\newblock Automated foosball table.
\newblock {\em California Polytechnic State University, Mech. Eng. Dep.}, 2013.

\bibitem{resnet}
Kaiming He, Xiangyu Zhang, Shaoqing Ren, and Jian Sun.
\newblock Deep residual learning for image recognition.
\newblock {\em arXiv}, 2015.

\bibitem{hernandez2019design}
Nicol{\'a}s Hern{\'a}ndez, Javier Rebolledo, Javier Torres, Gonzalo Carvajal, and Francisco Vargas.
\newblock Design of an experimental platform for the automation of the goalkeeper of a foosball table.
\newblock In {\em 2019 IEEE CHILECON}, pages 1--7. IEEE, 2019.

\bibitem{hintjens2013zeromq}
Pieter Hintjens.
\newblock {\em ZeroMQ: messaging for many applications}.
\newblock " O'Reilly Media, Inc.", 2013.

\bibitem{Horst2024}
Ronny Horst, David Hagens, Elke Hergenr{\"o}ther, and Andreas Weinmann.
\newblock Real time state detection of a foosball game using cnn-based computer vision.
\newblock {\em SAI Computing Conference, London (accepted)}, 2024.

\bibitem{mobilenet}
Andrew Howard, Mark Sandler, Grace Chu, Liang-Chieh Chen, Bo~Chen, Mingxing Tan, Weijun Wang, Yukun Zhu, Ruoming Pang, Vijay Vasudevan, Quoc~V. Le, and Hartwig Adam.
\newblock Searching for mobilenetv3.
\newblock {\em arXiv}, 2019.

\bibitem{janssen2010real}
Rob Janssen, Jeroen de~Best, and Ren{\'e} van~de Molengraft.
\newblock Real-time ball tracking in a semi-automated foosball table.
\newblock In {\em RoboCup 2009: Robot Soccer World Cup XIII 13}, pages 128--139. Springer, 2010.

\bibitem{janssen2012ball}
Rob Janssen, Mark Verrijt, Jeroen de~Best, and Ren{\'e} van~de Molengraft.
\newblock Ball localization and tracking in a highly dynamic table soccer environment.
\newblock {\em Mechatronics}, 22(4):503--514, 2012.

\bibitem{kingma2017adam}
Diederik~P. Kingma and Jimmy Ba.
\newblock Adam: A method for stochastic optimization, 2017.

\bibitem{Kitano1997}
Hiroaki Kitano, Minoru Asada, Yasuo Kuniyoshi, Itsuki Noda, Eiichi Osawa, and Hitoshi Matsubara.
\newblock Robocup: A challenge problem for ai.
\newblock {\em AI Magazine}, 18(1), 1997.

\bibitem{mohebi2022study}
Dani Mohebi.
\newblock The study of semi-automated foosball table.
\newblock {\em Tampere University of Applied Sciences, Degree Program in Mechanical Engineering}, 2022.

\bibitem{openai2019dota}
OpenAI, Christopher Berner, Greg Brockman, Brooke Chan, Vicki Cheung, Przemys\l aw~D\k{e}biak, Christy Dennison, David Farhi, Quirin Fischer, Shariq Hashme, Chris Hesse, Rafal J\'{o}zefowicz, Scott Gray, Catherine Olsson, Jakub Pachocki, Michael Petrov, Henrique~Pond\'{e} de~Oliveira~Pinto, Jonathan Raiman, Tim Salimans, Jeremy Schlatter, Jonas Schneider, Szymon Sidor, Ilya Sutskever, Jie Tang, Filip Wolski, and Susan Zhang.
\newblock Dota 2 with large scale deep reinforcement learning.
\newblock {\em arXiv}, 2019.

\bibitem{Pham2022}
Duc~An Pham and Trung~Nghia Pham.
\newblock Determination of a tilt angle for the automatic balancing system with the inertial measurement unit mpu6050.
\newblock In {\em Proceedings of the AMAS2021}, pages 349--354, Cham, 2022. Springer.

\bibitem{Rohrer2021}
Tobias Rohrer, Ludwig Samuel, Adriatik Gashi, Gunter Grieser, and Elke Hergenr{\"o}ther.
\newblock Foosball table goalkeeper automation using reinforcement learning.
\newblock {\em LWDA}, pages 173--182, 2021.

\bibitem{Silver2016}
David Silver, Aja Huang, Chris~J. Maddison, Arthur Guez, Laurent Sifre, George van~den Driessche, Julian Schrittwieser, Ioannis Antonoglou, Veda Panneershelvam, Marc Lanctot, Sander Dieleman, Dominik Grewe, John Nham, Nal Kalchbrenner, Ilya Sutskever, Timothy Lillicrap, Madeleine Leach, Koray Kavukcuoglu, Thore Graepel, and Demis Hassabis.
\newblock Mastering the game of go with deep neural networks and tree search.
\newblock {\em Nature}, 529(7587):484--489, Jan 2016.

\bibitem{stefani2014automated}
Jim~R Stefani, Alex~J Herpy, Brett~Gordon Jaeger, Kevin~S Haydon, and Derek~Alan Hamel.
\newblock Automated foosball table.
\newblock {\em California Polytechnic State University, Mech. Eng. Dep.}, 2014.

\bibitem{efficientnet}
Mingxing Tan and Quoc~V. Le.
\newblock Efficientnetv2: Smaller models and faster training.
\newblock {\em arXiv}, 2021.

\bibitem{weigel2005kiro}
Thilo Weigel.
\newblock Kiro - a table soccer robot ready for the market.
\newblock {\em Proceedings IEEE Int. Conf. Robotics and Automation}, 2005:4266--4271, 2005.

\bibitem{weigel2003kiro}
Thilo Weigel and Bernhard Nebel.
\newblock Kiro -- an autonomous table soccer player.
\newblock In G.~Kaminka, Pedro Lima, and Ra{\'u}l Rojas, editors, {\em RoboCup 2002: Robot Soccer World Cup VI}, pages 384--392. Springer, 2003.

\bibitem{weigel2004adaptive}
Thilo Weigel, Dapeng Zhang, Klaus Rechert, and Bernhard Nebel.
\newblock Adaptive vision for playing table soccer.
\newblock In Susanne Biundo, Thom Fr{\"u}hwirth, and G{\"u}nther Palm, editors, {\em KI 2004: Advances in {AI}}, pages 424--438. Springer, 2004.

\bibitem{zhang2007learning}
Dapeng Zhang and Bernhard Nebel.
\newblock Learning a table soccer robot a new action sequence by observing and imitating.
\newblock In {\em Proceedings of the Third AAAI Conference on Artificial Intelligence and Interactive Digital Entertainment}, volume~3, pages 61--66. AAAI Press, 2007.

\end{thebibliography}
\end{document}